\documentclass[letterpaper, 10 pt, conference]{ieeeconf}  

\IEEEoverridecommandlockouts                              

\usepackage{graphics} 
\usepackage{epsfig} 
\usepackage{mathptmx} 
\usepackage{times} 
\usepackage{amsmath} 
\usepackage{amssymb}  
\usepackage{amsfonts}
\usepackage{algorithm,algorithmic}
\usepackage{textcomp}
\usepackage{xcolor}
\usepackage{subfiles}
\usepackage{pdfsync}
\usepackage[hidelinks]{hyperref}
\usepackage{cite}
\usepackage{multirow}
\usepackage{makecell}

\title{\LARGE \bf
TDLE: 2-D LiDAR Exploration With Hierarchical Planning Using Regional Division
}

\author{Xuyang Zhao$^{*}$, Chengpu Yu, Erpei Xu and Yixuan Liu
    \thanks{* This work was supported by the National Natural Science Foundation of China (NSFC) (Grant Nos. 61991414 and 61873301), National Key Research and Development Project (2020YFC1512503) and Beijing Natural Science Foundation (L221005).}
    \thanks{All authors are with School of Automation, Beijing Institute of Technology, Beijing, China. 
    {\tt\small \{xuyang\_zhao, yuchengpu, erpei\_xu, yixuan\_liu \}@bit.edu.cn}}%
}

\begin{document}

\maketitle
\thispagestyle{empty}
\pagestyle{empty}

\begin{abstract}

Exploration systems are critical for enhancing the autonomy of robots.
Due to the unpredictability of the future planning space, existing methods either adopt an inefficient greedy strategy, or require a lot of resources to obtain a global solution.
In this work, we address the challenge of obtaining global exploration routes with minimal computing resources. 
A hierarchical planning framework dynamically divides the planning space into subregions and arranges their orders to provide global guidance for exploration.  
Indicators that are compatible with the subregion order are used to choose specific exploration targets, thereby considering estimates of spatial structure and extending the planning space to unknown regions.
Extensive simulations and field tests demonstrate the efficacy of our method in comparison to existing 2D LiDAR-based approaches.
Our code has been made public for further investigation\footnote{Available at \url{https://github.com/SeanZsya/tdle}}.

\end{abstract}


\section{Introduction}

Mobile robots equipped with 2-D LiDAR have found active applications in indoor cleaning, warehouse logistics, catering delivery, and many other scenarios.
However, traditional navigation algorithms for robots rely heavily on pre-built maps and manually set targets. To further improve the autonomy of these robots, exploration systems, which can guide robots to traverse the unknown scene and build environment map, have become a popular topic in robotics research.

After decades of study, various exploration methods for robots equipped with 2-D LiDAR have been proposed\cite{dudek1978robotic, yamauchi1997frontier,umari2017autonomous,chaplot2020learning}.
However, the existing methods still have limitations that hinder their practical application, as their low exploration efficiency, high computational overhead, or lack of autonomy. Those limitations are mainly because:

1) Most existing methods can't balance global perspective and computational efficiency. Some barely use greedy strategies that result in redundant route, while others use computationally expensive methods, like solving Travelling Salesman Problem (TSP) for every candidate point, or using learning-based methods to get prior information.

2) Inadequate indicators are selected for exploration revenue evaluation. For example, many of them use euclidean distance to target or the unknown area in Field of View (FoV) as indicators, which are rough estimations based on the current location and poorly reflects the benefits of actually reaching the target.



Motivated by the facts above, in this letter, we propose \textbf{TDLE}, an \textbf{2}-\textbf{D} \textbf{L}idar \textbf{E}xploration system with a hierarchical planning structure. It aims to provide a global perspective for exploration planning in an intuitive and efficient way. The proposed method divides map into different area, arranges the route of subregions firstly, then goes to the order of target points in each subregion.

Under this framework, after extracting frontier points, the mapping area is evenly divided into several subregions. The exploration route of subregions is determined by the route's similarity with the route before, the distance to initial grid and the total route length. 
Within each separated region, target points are then chosen using a revenue evaluation method that considers the distance to the adjoining edge, the number of visible frontier points, and the orientation difference.  


The proposed method is evaluated in both simulation and real-world environments, and the results demonstrate its superiority in exploration efficiency, calculation speed, and spatial traversal integrity. The contributions are summarized as follows:


1) A hierarchical planning strategy that employs regional division and arrangement to gain a global perspective for exploration effectively and efficiently.

2) A comprehensive revenue calculation method was developed to acquire targets that are compatible with the global plan while minimizing redundancy, without using information outside the current subregion.

3) An exploration system with full-process autonomy is built by redesigning or optimizing mapping, decision-making, and motion planning modules. The source code of the proposed system has been made public.


\section{Related Work}

Autonomous exploration system has been investigated for decades but remain unsolved. Unlike the coverage path planning problem\cite{galceran2013survey}, the unpredictability of future planning space makes it impossible to define the globally optimal route. Under this situation, the majority of existing systems rely on greedy decision-making approaches, such as selecting the nearest target or one that offers the most information gain among candidate points.

According to their difference in choosing candidate targets, conventional methods can be divided into frontier-based approaches and sampling-based approaches. Frontier-based approaches, such as \cite{yamauchi1997frontier, umari2017autonomous, cieslewski2017rapid, orvsulic2019efficient}, utilize frontier areas or points as exploration targets. On the other hand, sampling-based methods, like \cite{senarathne2016towards, bircher2016receding}, draw inspiration from the Next Best View (NBV) \cite{connolly1985determination} concept and generate random points in free space as potential targets near the current area.

A common issue of these methods is their lack of global perspective, resulting in low exploration efficiency.
To address this challenge, several methods \cite{zhou2021fuel,cao2021tare,petit2022tape} have been proposed to obtain a global planning route by solving a Travelling Salesman Problem (TSP) among candidate points. Some other methods utilize strategies like forward simulation\cite{lauri2016planning}, building skeleton graph\cite{dang2019graph} or optimal A* algorithm\cite{lee2021real} to exploit global knowledge. However, these methods demand significant computational resources and impose a heavy burden for edge computing devices.

On the other hand, learning-based methods, such as\cite{chaplot2020learning, bigazzi2022focus, niroui2019deep}, are able to provide prior knowledge of unknown area. These methods have potential in understanding the connection of spatial structures and make reasonable choices. However, they suffer from poor adaptability to unstructured environments, and the problem of high computing resources remains unsolved.

In this paper, we use regional division to obtain global exploration routes with minimal additional computing. Regional division have been used previously in some exploration methods\cite{jain2012multi, fermin2017tigre, alitappeh2022multi, zhou2023racer}, but they are either for multi-robot task allocation or limited in already known areas. We demonstrate its effectiveness in extending the planning space to unknown area and obtaining the global path.


\begin{figure}[t!]
    \centering
    \centerline{\includegraphics[width=0.95\linewidth]{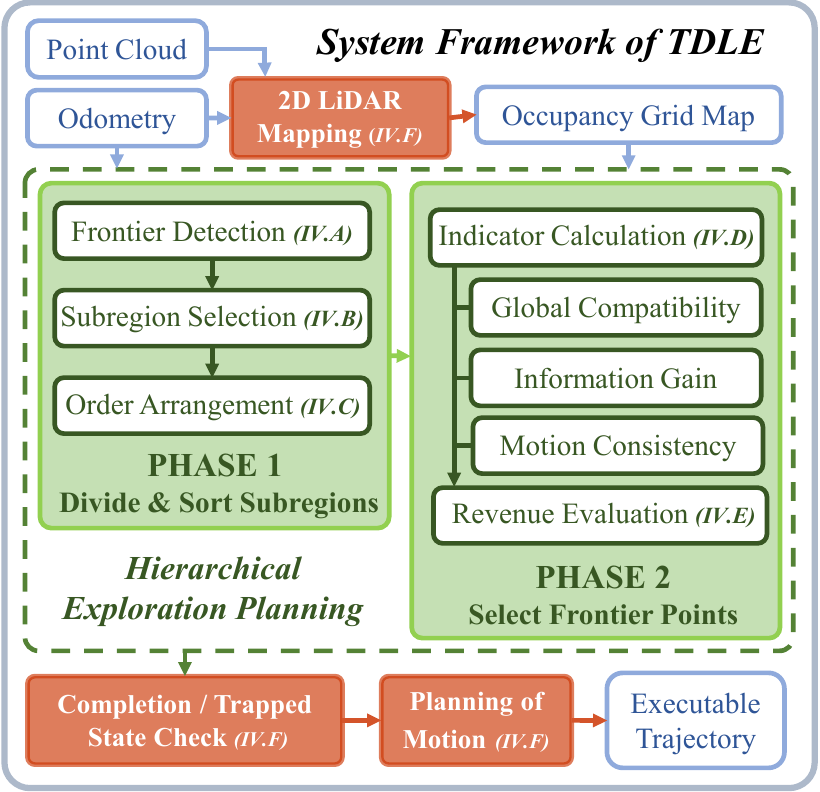}}
    \caption{Framework of The Proposed System}
    \label{fig:framework}
    \vspace{-2em}
\end{figure}


\section{System Overview}

As shown in Fig. \ref{fig:framework}, the proposed system take point cloud and odometry as input, and output an executable trajectory. The hierarchical exploration planning module is the core component of this framework.

In the first phase, the planning space is divided into several subregions (Sect.\ref{sec:regional-division}). After that, subregions with frontier points inside or mostly unknown are selected (Sect.\ref{sec:selection-of-subregions}). The selected subregions are then sorted to provide global guidance for exploration (Sect.\ref{sec:subregion-order-arrangement}).

In the second phase, the subregions are visited one by one in the order determined in the first phase. In each subregion, various indicators, including global compatibility, information gain, and motion consistency, are calculated (Sect.\ref{sec:indicators-calculation}). Robot selects the exploration target with the highest comprehensive revenue (Sect.\ref{sec:revenue-evaluation}).

Additionally, the framework integrates supporting modules that connect the entire system, including 2D LiDAR mapping (Sect.\ref{sec:2d-lidar-mapping}), completion and trapped check, and motion planning (Sect.\ref{sec:motion-planning}). These modules help to further increase the autonomy and efficiency of the proposed system.

\section{Proposed Method}

\subsection{Frontier Points Detection}

    Like many other exploration methods\cite{umari2017autonomous,bircher2016receding}, we use Rapidly-exploring Random Trees (RRT) to detect frontier points, for its several advantages like fast running speed and probabilistic completeness.

    The frontier searching module is adapted from \cite{umari2017autonomous}, with the following adaptations to reduce computational overhead and enhance autonomy: 

    1) The global tree is limited to only $n_{nd}$ nodes, where $n_{nd}$ is dynamically adjusted with the map size to strike a balance between computational efficiency and search integrity.
        
    2) The sampling space is now expanding automatically to match the size of the current map, instead of manually selecting, thus avoiding sampling in extra space.

    We further eliminate a candidate point if it has too few unknown grids within a small radius $\varepsilon$, or it's close to other candidates. Finally, the set ${F\!P}$ of frontier points $p_i$ is generated, where ${F\!P} = \{p_0, p_1,..., p_n\}, n \leq n_{fp}$.

\subsection{Subregion Selection}

    After frontiers are detected, a hierarchical planning framework will decide the travelling order of every frontier point. As in Fig. \ref{fig:framework}, the first phase of this framework is to handle subregions with a divide-select-arrange pattern, aiming to provide a global perspective to the exploration.

    \subsubsection{Regional Division}
    \label{sec:regional-division}

    To reduce the computational complexity, we do not perform a specific analysis of the structure of the space.
    Instead, the Axis-Aligned Bounding Box (AABB) of current map is used for division.
    
    As demonstrated in Algorithm.\ref{alg: sort_sr}, the AABB with size of $l_{box} \times h_{box}$ is evenly divided into subareas $SR_{all} = \{sr^{all}_0,sr^{all}_1,...,sr^{all}_n\}, n<n_l n_h $, where $n_l$ and $n_h$ are the numbers of subregions in the horizontal and vertical directions, respectively.

        \begin{algorithm}[t]
        \caption{Subregion Arrangement}
        \label{alg: sort_sr}
        \begin{algorithmic}[1]
            \renewcommand{\algorithmicrequire}{\textbf{Input:}}
            \renewcommand{\algorithmicensure}{\textbf{Output:}}
            \REQUIRE{$SR_{sel}$: list of subregions, $\eta$: initial cooling rate, $\mu$: decay rate, $T_0$: initial temperature, $T_{stop}$: stopping temperature, $n_{ite}$: maximum number of iterations}
            \ENSURE{$R_{opt}$: optimal arrangement}
            \STATE $R_0 \gets \text{Route that visit all } sr_i \text{ from } sr_0 \text{ randomly}$
            \STATE $\text{Set } R_{opt} \gets R_0$
            \STATE $\text{Set } T \gets T_0$
            \WHILE{$T > T_{stop}$ and $n_i \leq n_{ite}$}
            \STATE $R_i \gets R_{i-1} \text{ with two elements swapped} $
            \STATE $\Delta C_{temp} \gets C_{temp}(R_i) - C_{temp}(R_{opt})$
            \IF{$\Delta C_{temp} \geq 0$}
            \STATE $R_{opt} \gets R_i$
            \ELSE
            \STATE $R_{opt} \gets R_i \text{, by probability } P = e^{-\Delta C_{temp}/T}$
            \ENDIF
            \STATE $ \eta \gets \eta =  e^{ \mu (\frac{n_i}{n_{ite}} -1)} \text{\quad // Update cooling rate}$ 
            \STATE $ T \gets T \times \eta \text{\quad // Decrease temperature}$
            \STATE $n_i \gets n_i + 1$
            \ENDWHILE
            \RETURN $R_{opt}$
        \end{algorithmic}
    \end{algorithm}

    The segmentation process is initialized with $3 \times 3$ grids. Once the length or height of the subregion is more than twice the diameter $d_{lid}$ of LiDAR's Field of View (FoV), we determine that the subregion is no longer sufficient to represent the local properties, and the corresponding value of $n_{l}$ or $n_{h}$ is increased by one.

    
    \subsubsection{Selection of Subregions}
    \label{sec:selection-of-subregions}

    After the area decomposition, we consider subregions with frontier points inside or mostly unexplored as areas worth exploring. To define if a subregion is mostly unknown, function $isU\!nknown(sr^{all}_i)$ samples points in $sr^{all}_i$ and check if the ratio of unknown points exceeds a certain threshold. 
    
    Finally,  $n_{sr}$ subregions are selected for further arranging, forming set $SR_{sel} = \{sr^{sel}_0, sr^{sel}_1,..., sr^{sel}_n\}, n \leq n_{sr}$. The subregion $sr_0$ corresponds to the robot's current location $p_{bot}$.

\subsection{Subregion Order Arrangement}
\label{sec:subregion-order-arrangement}

\begin{figure}[t]
        \centering
        \centerline{\includegraphics[width=1\linewidth]{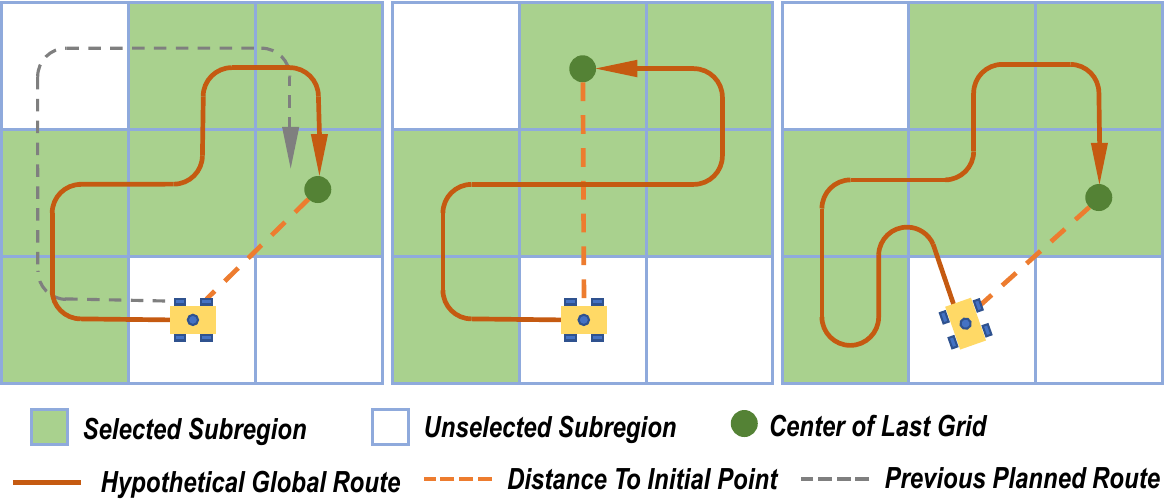}}
        \caption{Possible Global Routes. From left to right: ideal global route, with last grid far from initial position, having redundant segments. The middle and the right route also have lower similarity to previous planned route.}
        \label{fig:sort_sr}
        \vspace{-3em}
    \end{figure}

    
    In the following steps, $SR_{sel}$ is sorted to create an outline of the global route for exploration. As show in Fig. \ref{fig:sort_sr}, to ensure exploration efficiency and time consistency of planning, an ideal route should have the following properties:

    \begin{itemize}
        \item Starting from the subregion $sr^{sel}_0$;
        \item Visiting every subregion in $SR$ at least once;
        \item With total length as short as possible;
        \item Similar to the previous planned route;
        \item End with the last subregion close to initial point.
    \end{itemize} 
    
    To satisfy those conditions, we employ an Adaptive Simulated Annealing (ASA) algorithm to generate the global route, as in Algorithm.\ref{alg: sort_sr}. Denote the route consisting the center point of subregions in $SR_{sel}$ as an ordered list $R_i = [r^i_0, r^i_1, ..., r^i_j], j \leq n_{sr}$. Initializing the solution $R_{opt}$ with a route $R_0$ that starts from $sr_0$ and visits all $sr_i$ in a random order. 
    
    Set the initial temperature as $T_0$ and the cooling rate as $\eta$. Repeat the following iteration until the temperature reaches a certain threshold $T_{stop}$ or a maximum number $n_{ite}$ of iterations is reached. For the $n_i$-th iteration:

    New route $R_i$ is generated by randomly swapping two elements in $R_{i-1}$, then the score $C_{temp}(R_i, R_{opt})$ is calculated by the summary of similarity $C_{sim}(R_i)$ between $R_{i}$ and $R_{opt}$, the distance to initial grid and the total length: 
    \begin{equation}
        \begin{aligned}
        C_{temp}(R_i, R_{opt}) & = \lambda_{s}C_{sim}(R_i,R_{opt}) - \\
        & \lambda _{d}\|\mathbf{r^i_{n_{sr}}} - \mathbf{p_{ini}} \| - \lambda_{l}\sum_{j=1}^{n_{sr}}\|\mathbf{r^i_j} - \mathbf{r^i_{j-1}}\|
        \end{aligned}
    \label{equ:c_temp}
    \end{equation}

    $C_{temp}(R_i, R_{opt})  = \lambda_{s}C_{sim}(R_i,R_{opt}) -  \lambda _{d}\|r^i_{n_{sr}} - p_{ini} \| - \lambda_{l}\sum_{j=1}^{n_{sr}}\|r^i_j - r^i_{j-1}\|$
    where $\lambda_{d}$, $\lambda_{l}$ and $\lambda_{s}$ are the tuning factors. The similarity $C_{sim}(R_i,R_{opt})$ between routes $R_i$ and $R_{opt}$ is evaluated by Dynamic Time Warping (DTW) algorithm \cite{keogh2005exact}. It generates an accumulated cost matrix $M$ and find the best path by backtracking from the upper right corner of it. An element $M(a,b)$ in $M$ is calculated by:
    \begin{equation}
        \begin{aligned}
        M(a,b) & =  \| \mathbf{r^{opt}_a} - \mathbf{r^{i}_b} \| + min \{ M(a-1,b-1),\\
        &  M(a, b-1), M(a-1,b) \}
        \end{aligned}
    \end{equation}
    
    If the new route $R_i$ is better ($C_{temp}(R_i, R_{opt}) \geq 0$), accept it as new $R_{opt}$. If it's worse, accept it by a probability $P = e^{-{\Delta C_{temp}}/T}$ to avoids tucking in local optimum. Then the temperature is updated by $T = \eta T$. To accelerate the search process, $\eta$ is updated adaptively using $\eta =  e^ { \mu (n_i/n_{ite} -1)}$, where $\mu$ is the decay rate.
    

      
    When the loop ends, the current optimal solution is selected as solution $R_{opt}$. The final list of subregions $SR_{opt} = [sr^{opt}_0, sr^{opt}_1, ..., sr^{opt}_i], i < n_{sr}$ is obtained by replacing center point $r^{opt}_j$ with its corresponding subregion $sr^{opt}_i$. 
    
\begin{figure*}[t]
        \centering
        \includegraphics[width=1\linewidth]{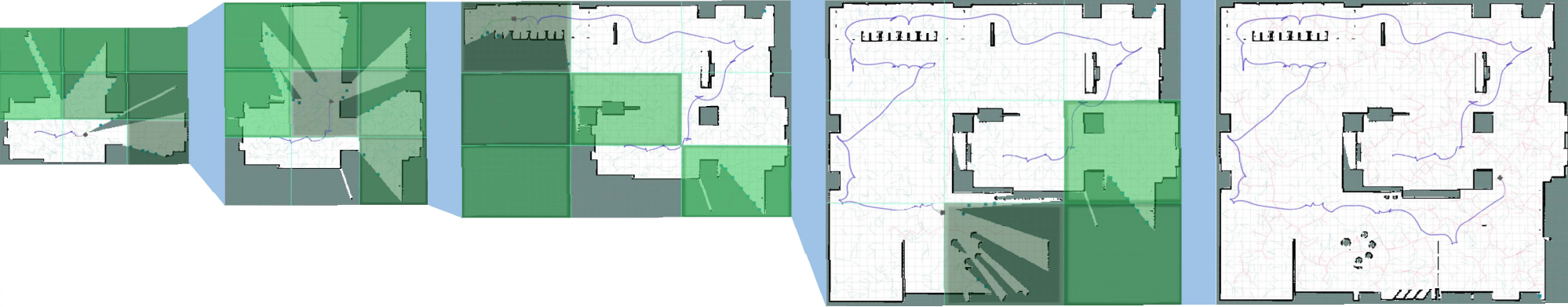}
        \caption{The exploration process of the proposed method. Images from left to right reflect the different periods of exploration. The color of the subregions represents the order of visiting them in global planning (from darker green to brighter green). The subregions are updated with the map size and the previous map becomes a subregion of the later division.}
        \label{fig:exploring_sim}
        \vspace{-2em}
      \end{figure*}

    To be noticed, as depicted in Fig. \ref{fig:exploring_sim}, the previous planning space becomes a subregion within the subsequent space, thus early routes will not conflict with later ones.

\subsection{Indicator Calculation}
\label{sec:indicators-calculation}

    The sequence $SR_{opt}$ for traversing the subregions is determined in the first phase. In the second phase, the frontier point with the highest exploration revenue in each subregion is selected as the specific exploration target.
    
\begin{figure}[h!]
        \centering
        \centerline{\includegraphics[width=\linewidth]{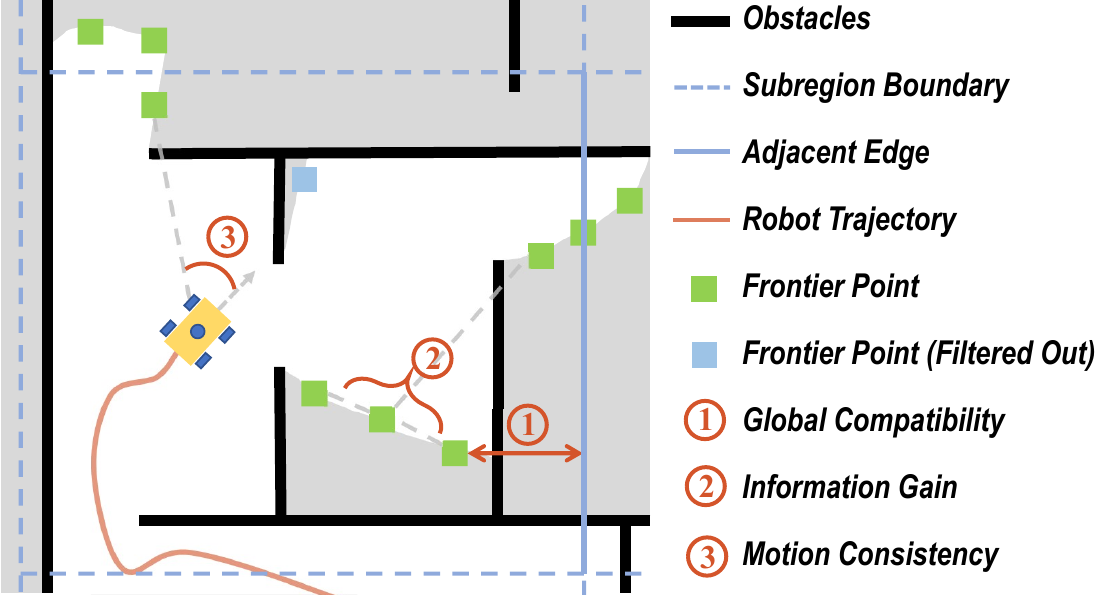}}
        \caption{Indicators For Revenue Calculation}
        \label{fig:indicators}
    \end{figure}

    When selecting the indicators for the calculation of exploration revenue, the following criteria were taken into account for a more effective and streamlined exploration trajectory:
    \begin{itemize}
        \item Compatibility with the regional sequence;
        \item Efficient reflection of obtained information;
        \item Ability to help reduce excess trajectory.
    \end{itemize}
    As show in Fig. \ref{fig:indicators}, global compatibility, information gain and motion consistency are chosen as indicators. The following subsections elaborate on each indicator.

    \subsubsection{Global Compatibility}

    We adopt global compatibility $G_{com}(p_i)$ instead of the conventional navigation cost indicator. $G_{com}(p_i)$ describes how well a frontier point $p_i$ aligns with the global route. It is computed as the distance between $p_i$ and the adjoining edge. 

    Ideally, the robot would move to the adjoining edge from far to near, and eventually to the next subregion $SR_{i+1}$. In this way, it can directly utilize the outcomes of the subregion assignment and obtain a global perspective without considering information beyond the current subregion $SR_i$. Furthermore, it's readily calculated with linear time complexity $O(n)$.

    \subsubsection{Information Gain}

    The information gain $G_{inf}(p_i)$ measures the amount of environment information that can be obtained by reaching $p_i$. Estimating $G_{inf}(p_i)$ can be computationally expensive in various systems \cite{schmid2020efficient} as they search the entire field of view (FoV) of $p_i$ to count the number of unknown grids. 
    
    In the proposed system, $G_{inf}(p_i)$ is defined as the number of frontier points $p_j$ in $F\!P$ that can be observed from $p_i$. As the candidate points are generated uniformly, $p_i$ in $F\!P$ tend to be evenly distributed across the frontier area. Hence, $G_{inf}(p_i)$ can effectively reflect the size of the detectable unknown area, particularly in wider frontier areas such as doors and corridor corners.

    \subsubsection{Motion Consistency}
    One crucial approach to enhance exploration efficiency is by reducing redundant paths. 
    Although this consideration has been taken by former indicators, it's still necessary to use motion consistency to directly minimize zigzag routes.

    Denote the angle difference between the robot's orientation and the line connecting $p_{bot}$ and $p_i$ as $\alpha_{ori}$, the motion consistency $C_{mot}(\alpha _{ori})$ is defined as follows:
    \begin{equation}
        C_{mot}(\alpha _{ori})= e ^{ 2\times(\frac{2 \alpha_{ori}}{\pi} - 1)}
    \end{equation}

    By employing this formulation, larger values of $\alpha_{ori}$ can be penalized more significantly since they directly contribute to unnecessary routes. While preventing small values of $\alpha_{ori}(p_i, p_{bot})$ from being the decisive factor, as the motion planning algorithm can generate routes that facilitate gradual orientation changes.


\subsection{Comprehensive Revenue Evaluation}
\label{sec:revenue-evaluation}

    To standardize the data scale before revenue evaluation, the indicators of $p_i$ are normalized with other candidate points in the same subregion $sr^{opt}_i$ using z-score normalization. Subsequently, the next target $p{tgt}$ is determined as the one that maximizes the comprehensive revenue  $R_{ev}(p_i)$:
    \begin{gather}
        p_{tgt} = \arg \max _{p_i \in FP}  R_{ev}(p_i) \\
        R_{ev}(p_i) = \lambda_c G_{com}(p_i)  + \lambda_i G_{inf}(p_i) - \lambda_m  C_{mot}(\alpha_{ori}) 
        \label{equ:r_ev}
    \end{gather}
    where $\lambda_c$, $\lambda_i$ and $\lambda_m$ are the weights of three indicators. Finally, $p_{tgt}$ will be assigned to motion planning module for further exploration.

\subsection{Supporting Modules}
    Autonomous exploration is a complex problem that integrates multiple fields. Mapping and motion planning, while not directly affecting exploration decision-making, are essential for keeping the system functioning.

    \subsubsection{2D LiDAR Mapping}
    \label{sec:2d-lidar-mapping}
    Cartographer\cite{hess2016real} is used to generate the occupancy grid map for its high accuracy when compared to other mapping algorithms\cite{yagfarov2018map}. To make it compatible with the exploration system, we modified its message conversion module to generate binarized occupancy grid map.

    \subsubsection{Motion Planning}
    \label{sec:motion-planning}
    
    In motion planning, the selected point $p_{tgt}$ will be assigned as the next target. A* and Timed Elastic Band (TEB)\cite{rosmann2017kinodynamic} algorithms are then employed to generate an obstacle-free and executable path. Additionally, we implement a state check module to enhance autonomy. Once the module notices that the robot has stopped, it verifies if $F\!P$ is empty. If it is, the exploration is considered complete, and $p_{ini}$ is sent for autonomous return. If it's not, the robot may be trapped by obstacles, and former waypoints will be sent to guide the robot to escape.

\begin{figure}[t]
  \includegraphics[width=\linewidth]{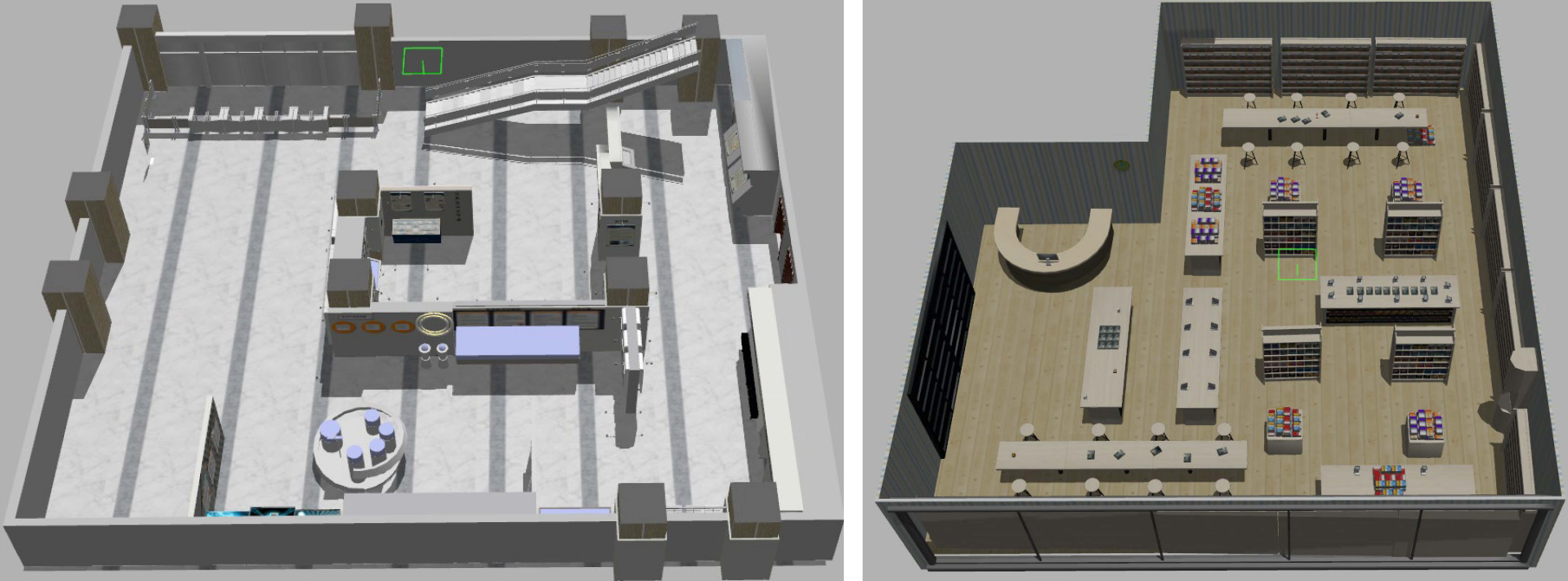}
  \caption{Simulated environments in Gazebo. Left: museum; right: library.}
  \label{fig:time_alc}
  \vspace{-4em}
\end{figure}

\section{Experiments}





\subsection{Simulation Experiments}

The proposed method is benchmarked with two 2D LiDAR exploration methods: the frontier-based method with dense frontier detection\cite{orvsulic2019efficient} and the multi-RRT exploration\cite{umari2017autonomous}. The simulation tests are conducted in Gazebo simulator, and runs on a 1.4GHz Intel Core i5-10210U CPU with 16GB RAM. 

We select two typical application scenarios for service robots as the simulation experiments: a museum and a library. The museum is a large ($ 450 m^2 $) open space with a looped corridor, while the library is comparatively small ($250 m^2$) and spread with many obstacles. In each scenario, simulation experiments were conducted 10 times with each method.

In our approach, we do not use the ratio of explored area to time, but the ratio of explored area to distance traveled as a measure of system efficiency. For the time spent on exploration is influenced not only by the rationality of exploration planning, but also by other factors like the robot's motion performance, making it difficult to compare results across different platforms.



\begin{table}[h!]
  \centering
   \caption{Exploration statistic in the museum and library scenarios}
   \label{tab:exp_stat}
  \begin{tabular}{p{0.4cm}<{\centering}cp{0.5cm}<{\centering}p{0.4cm}<{\centering}p{0.4cm}<{\centering}p{0.4cm}<{\centering}c}
    \hline\hline
  \multirow{2}{*}{\textbf{Scene}}  & \multirow{2}{*}{\textbf{Method}} & \multicolumn{4}{c}{\textbf{Travel Distance (m)}} &  \multicolumn{1}{c}{\multirow{2}{*}{\textbf{\begin{tabular}[c]{@{}c@{}}Exploration\\ Rate\end{tabular}}}} \\ \cline{3-6}
    &      & Max      & Min     & Std       & Avg        &   \\ \hline
  \multicolumn{1}{c}{\multirow{3}{*}{\begin{tabular}[c]{@{}c@{}}Museum\\ ($450m^2$)\end{tabular}}} 
  & Frontier    & 97.26   & 75.32   & 6.85    & 88.28   & 5.10\\
  & Multi-RRT   & 160.63  & 88.51   & 24.04   & 119.15  & 3.78  \\
  & Proposed    & 90.01   & 68.13   & 7.39    & \textbf{75.45}   & \textbf{5.96} \\ \hline
  \multirow{3}{*}{\begin{tabular}[c]{@{}l@{}}Library\\ ($250m^2$)\end{tabular}}
  & Frontier    & 85.00   & 75.77   & 3.93    & 81.17   & 3.08 \\
  & Multi-RRT   & 87.53   & 60.39   & 10.03   & 80.03   & 3.12 \\
  & Proposed    & 83.43   & 55.74   & 12.50   & \textbf{65.82}   & \textbf{3.79} \\
    \hline\hline
  \end{tabular}
  \end{table}

\begin{figure}[t]
  \includegraphics[width=\linewidth]{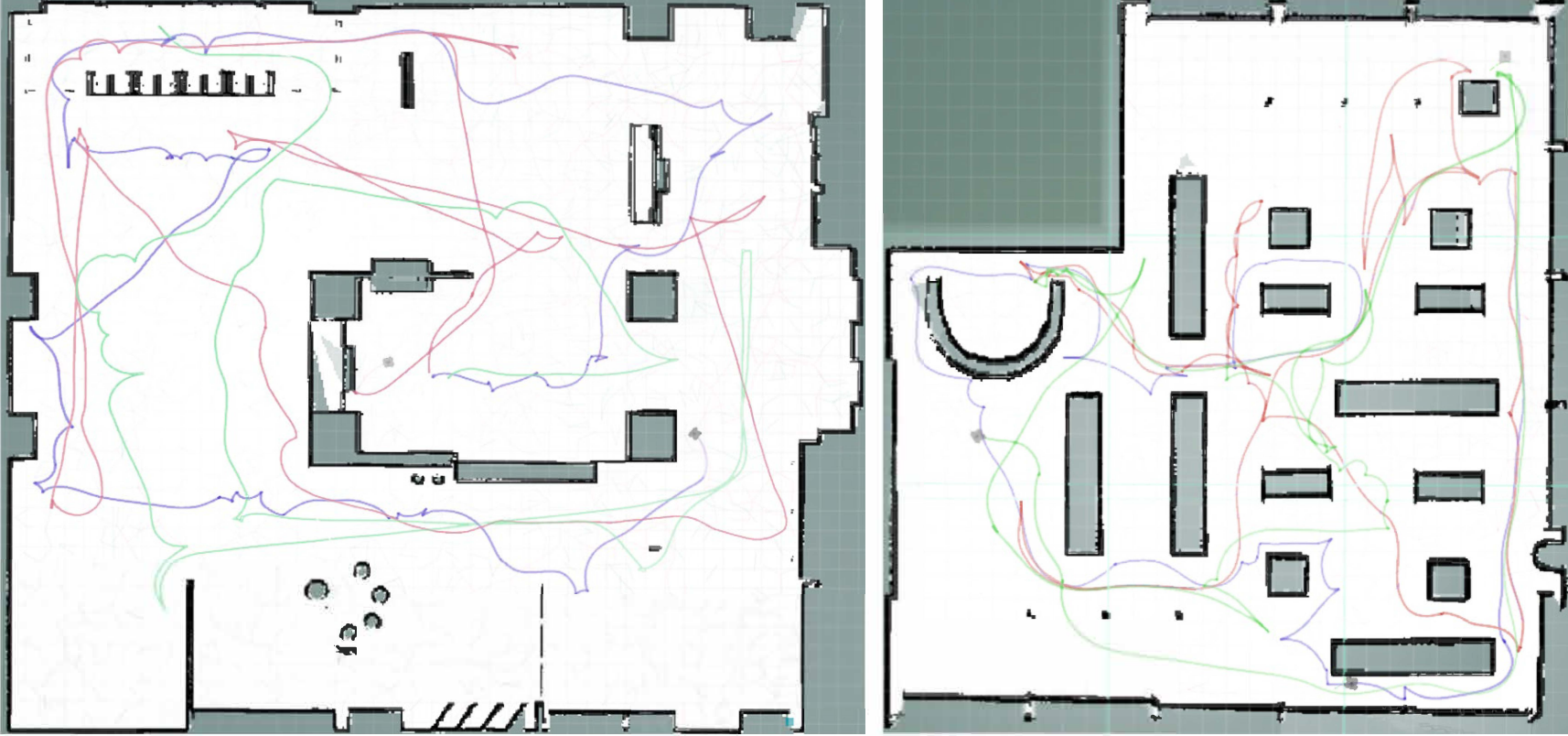}
\caption{Exploration results and trajectories. The built maps of museum (left) and library (right) are visualized, together with trajectories of frontier method (green), multi-RRT (red), and proposed method (purple).}
\label{fig:traj}
\end{figure}

\begin{figure}[t]
  \includegraphics[width=\linewidth]{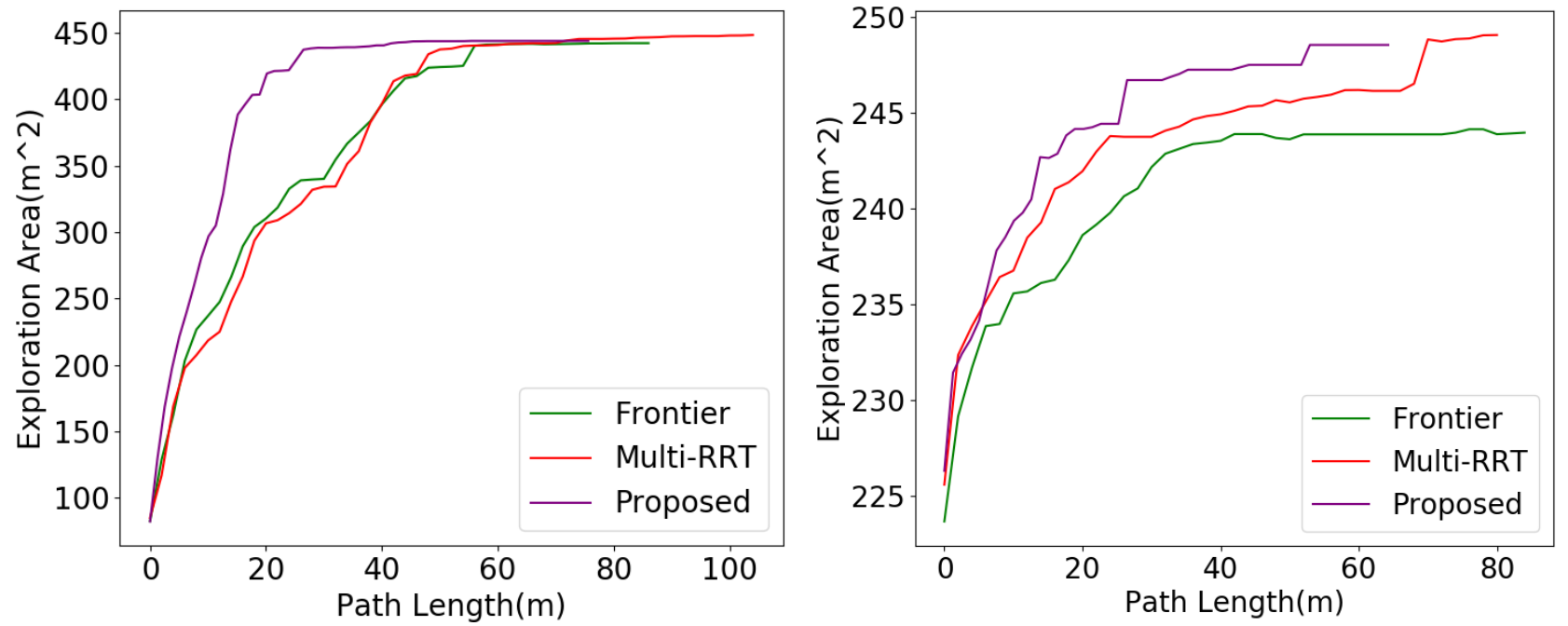}
\caption{The average value of the exploration area and the corresponding path length in scenarios of museum (left) and library (right).}
\label{fig:rate}
\vspace{-2em}
\end{figure}

Table.\ref{tab:exp_stat} presents the experimental results and statistics, and the exploration result and trajectories can be found in Fig. \ref{fig:traj}. The average value of the exploration area and the corresponding path length is shown in  Fig. \ref{fig:rate}. 

The results indicate that the hierarchical planning strategy effectively reduces the generation of redundant paths. Our method can use shorter paths to obtain more information of the unknown environment, and the exploration efficiency is improved by  $16.86\%$ and $57.67\%$ compared to Frontier and multi-RRT methods in museum scenario, and by $22.40\%$ and $21.47\%$ in library scenario.

We also compared the average computation time for global plan generation between different methods. Two other methods were used during the exploration process: (1) The Lin-Kernighan-Helsgaun (LKH) solver\cite{helsgaun2000effective}, implemented in \cite{zhou2021fuel,cao2021tare}, which sorts frontier points as a TSP problem; (2) TIGRE\cite{fermin2017tigre}, which utilizes contour-based segmentation and topological map searching. The comparation result is shown in Table.\ref{tab:time_consum}.

\begin{table}[h!]
  \centering
   \caption{Time consumption for global plan generating}
   \label{tab:time_consum}
  \begin{tabular}{p{2cm}<{\centering}p{1.5cm}<{\centering}p{1.5cm}<{\centering}}
    \hline\hline
  \multirow{2}{*}{\textbf{Method}} & \multicolumn{2}{c}{\textbf{Computation Time (ms)}} \\ \cline{2-3}
                                   & \textbf{Museum}         & \textbf{Library}         \\ \hline
  TSP-based                        & 126.62                  & 50.78                   \\
  TIGRE                             & 139.33                  &  215.57 \\
  Proposed                         & \textbf{7.07}          & \textbf{8.85}            \\
  \hline\hline
  \end{tabular}
\end{table}

The results indicate that the speed of obtaining the global plan has been significantly improved. The TSP problem has a time complexity of O(${2^n}{n^2}$), which significantly increases with the number of candidate points. While the contour-based segmentation in TIGRE is also computationally resource intensive for its using of dual-space decomposition. In contrast, the proposed method uses ASA to sort the subregions, whose time complexity is exactly limited by $n_{ite}$. Additionally, DTW is used for comparing similarity, with time complexity of O($n_1 \times n_2$). Given the small number of subregions, it also has a fast computation speed.

\subsection{Real-World Tests}
For the real-world experiments, we use a mobile robot equipping a 2D LiDAR with range of $10m$. The system runs on Nvidia Jetson Nano, an edge computing platform with a Cortex-A57 CPU and memory of 4GB. A maze structure is decorated to test its exploration ability.

\begin{figure}[t]
  \centering
  \centerline{\includegraphics[width=0.95\linewidth]{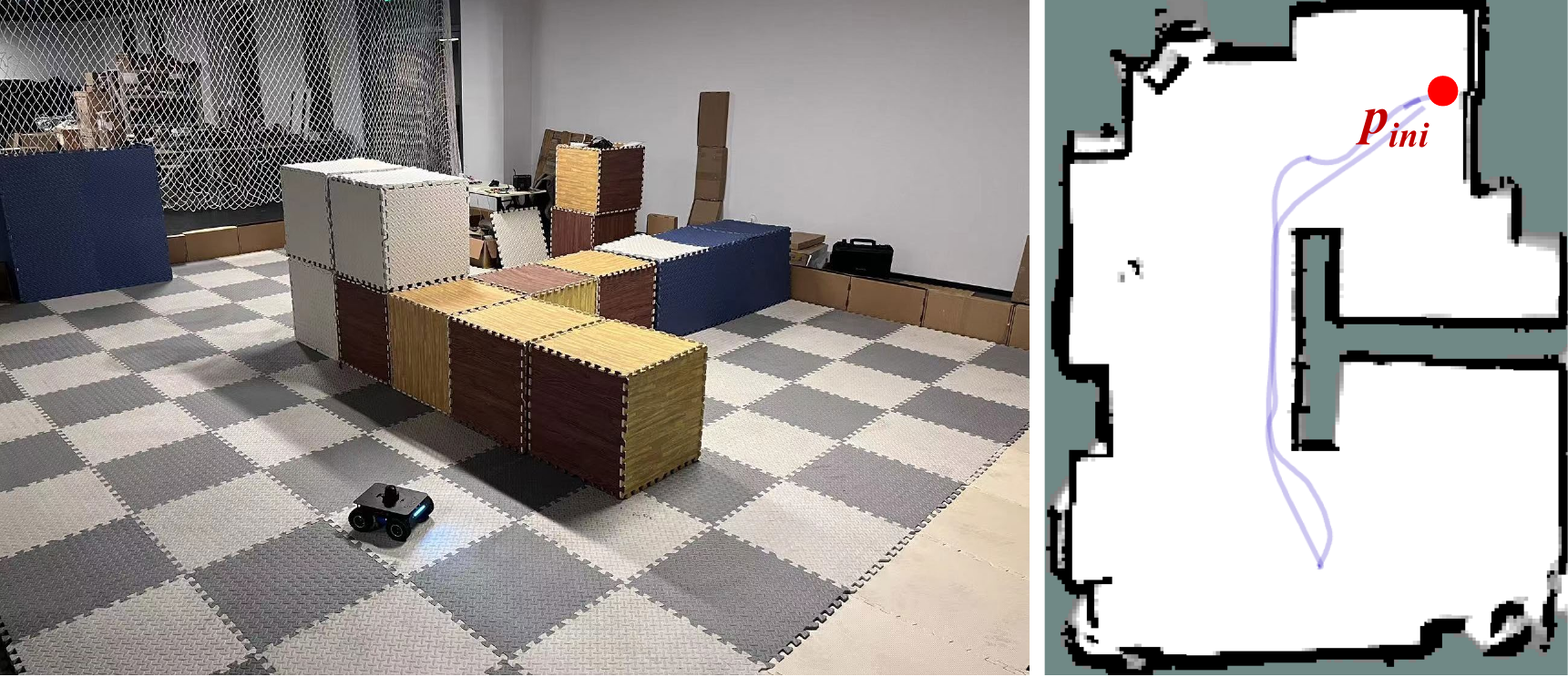}}
  \caption{Field test environment (left) and exploration result (right).}
  \label{fig:field_test}
  \vspace{-2em}
\end{figure}

As shown in Fig. \ref{fig:field_test}, the robot can successfully generate a global plan, explore the environment and return to initial point automatically. The results demonstrate that the proposed system is capable of handling realistic environments with a low cost and low computing power platform. For more information, please refer to the demonstration video on: \url{https://youtu.be/aPXxOKf1o10}.

\section{Conclusions}
\label{sec:conclude}

In this paper, a hierarchical planning framework has been proposed for obtaining global exploration routes in an intuitive and efficient way. The planning space has been dynamically divided into subregions and arrange their orders to provide global guidance for exploration. Indicators that compatible with the subregion order have been selected to choose specific exploration targets. Mapping and motion planning modules have also been optimized to further enhance the autonomy and efficiency of the proposed system. Extensive simulation and field tests have been conducted, demonstrating the effectiveness of our proposed method.


\addtolength{\textheight}{0.cm}  




\bibliographystyle{IEEEtran}
\bibliography{tdle}

\end{document}